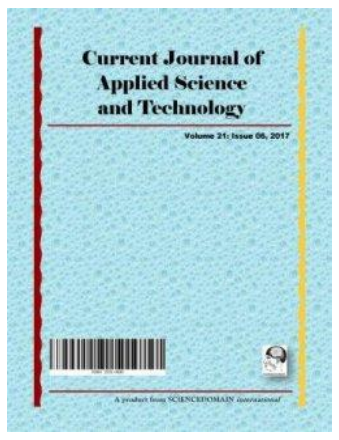

# Machine-Translation History and Evolution: Survey for Arabic-English Translations

**Nabeel T. Alsohybe[1*], Neama Abdulaziz Dahan[2] and Fadl Mutaher Ba-Alwi[3]**

*[1]Department of Information Technology, Sana'a University, Sana'a, Yemen.*
*[2]Department of Computer Science, Sana'a University, Sana'a, Yemen.*
*[3]Department of Information Systems, Sana'a University, Sana'a, Yemen.*

***Authors' contributions***

*This work was carried out in collaboration between all authors. Author NTA designed the study and wrote the first draft of the manuscript. Authors NAD and FMBA managed the analyses of the study, collected the data and performed the statistical analysis. Author FMBA managed the literature searches. All authors read and approved the final manuscript.*

## ABSTRACT

As a result of the rapid changes in information and communication technology (ICT), the world has become a small village where people from all over the world connect with each other in dialogue and communication via the Internet. Also, communications have become a daily routine activity due to the new globalization where companies and even universities become global residing cross countries' borders. As a result, translation becomes a needed activity in this connected world. ICT made it possible to have a student in one country take a course or even a degree from a different country anytime anywhere easily. The resulted communication still needs a language as a means that helps the receiver understands the contents of the sent message. People need an automated translation application because human translators are hard to find all the times, and the human translations are very expensive comparing to the translations automated process. Several types of research describe the electronic process of the Machine-Translation. In this paper, the authors are going to study some of these previous researches, and they will explore some of the needed tools

**Corresponding author: E-mail: alsohybe@gmail.com;*
*Coauthor Email: neama.abdulaziz@gmail.com, dr.fadlbaalwi@gmail.com;*

for the Machine-Translation. This research is going to contribute to the Machine-Translation area by helping future researchers to have a summary for the Machine-Translation groups of research and to let lights on the importance of the translation mechanism.

## 1. INTRODUCTION

Information and Communication Technology (ICT) has rapidly grown in the last decades. This affected business development [1] and the systems that depend on telecommunications where people can communicate with many other people from different societies. Thus, language is an essential communication means depending on the importance of the deals with other people from other communities and cultures [2] because of trading, studying, or friendship. People need languages to send messages and understand the messages they receive, too. Thus, it is important to have a translation method that is like exchanging two messages that are equivalent in two different codes [3].

The translation has been needed since the far away past [2] when human started having journeys for trading, and then for studying. The various deals with other people from different cultures that have different languages made it necessary for the government, at that time, to send some chosen individuals to learn those other people's language and culture. This way of interaction was how the human translation started. The human translator must have a huge experience, in that language before starting translating it. The translation itself needs substantial modeling of knowledge about the world, the translation sciences and the languages we translate between them [4]. The less the experience, the worse the translation [5]. The most important aspects of the human translation are that they are precise and always bilingual. However, the human translator does not exist every time we need them, and it is not always for free. Also, the privacy of the conversations or the translated messages depends on the translator's ethics and morals.

Nowadays, Translation is a significant activity in people' daily lives [6], and it is as old as the idea of the computer [7], as a result of the Telecommunication Technology which makes the communication much easier, anytime, anywhere and many times for free. Thus, the Computer Science engineers started thinking about making the machine able to translate messages which first introduced to the machine as texts. Unfortunately, translation is a difficult task [8] because of its dependency on the Natural Language Processing (NLP) and the structure of the languages [7]. The only way to make sense was by using the artificial intelligence because it merges the different required types of sciences during the translation process. A new field of the Machine Learning was to develop algorithms for the Natural Language Processing (NLP) [9]. These algorithms can be used to analyze the source texts [9] into its Parts-Of-Speech (POS) tags to facilitate the translation mechanism. These algorithms are Data Mining (DM) algorithms, such as the POS tools [10] which are the encoder or decoder in the Neural Machine Translation (NMT) [11]. Also, merged with the Data Mining (DM) algorithms in order to help during translation, such as the parsers or Morphological Analyzers tools those used in the Statistical Machine Translation (SMT) [12,13].

Many types of research are proposing Translations with the help of the machine. Machine Translation is a contemporary research field which means a translation that is providing a fully automated translation. It is different from the computer-aided translation, machine-aided human translation (MAHT) and the human-aided machine translation (HAMT) [14]. Computer-aided Translation (CAT) is a form of language translation in which a human translator uses computer software to support and facilitate the translation process [15]. MAHT is a human translation with the assistance of the machine. In this machine, there's a program that can check spelling, grammar, or the style of the translation [16]. HAMT is a machine that is responsible for translation process with an assistance of the human whenever needed. This human either can be involved while the process "interactive" or can be outside the process "pre-edit" or "post-edit" [16].

Many pieces of research discuss the Machine Translation, which comes in the following three categories: Statistical Machine Translation (SMT), Machine Translation using Semantic

Web, and Neural Machine Translation (NMT). To the best of our knowledge, none of these machines give precise translation like the human translation [17]. Except for the translation of the texts, this recorded in the knowledge base of the machine as assigned phrases which are the full phrases from the source language and its meaning in the target language [13,18]. That is because the words translation has been saved as it is to the knowledge base as translation experts translate it. However, we cannot insert every single phrase and ask the user to translate it. The wheel of the sciences did not stop. The Machine Translations using Semantic Web brings us a new chance for translation using the semantic meanings of the texts [12,19]. However, it is slower than other machines types [8], and if the meanings of the homograph words are not added to the ontology file, the translation could be bad.

Some other tools are developed according to much research to help in the Machine Translations area. Some examples of these tools are the Syntax Analyzers such as the parsers [20,21] and the Morphological analyzers [22-24]. These tools are especially used to analyze the text to tell the user, either the human or the Machine, the POS tags [10] and define the stems of the words. Some other machines build their Syntax analyzers such as the encoder of the NMT.

This research is going to provide a summary of the translation itself. Then the research will explore the different types of the Machine-Translation. Finally, the research will go over some needed tools while for Machine-Translation before the conclusion.

## 2. TRANSLATION

Translation is a very rapid growing activity these days [6]. It is an active means to transfer the culture and language due to the contact with one other [2]. Its main aspect understands the meaning assigned to the words in the vocabularies of the language [3]. Any translated text is judged and accepted by the reviewers when it is read as it was meant in its source language, in other words, it is not literally translated, this is because the author may express his or her thoughts and feelings in writing or speaking [25].

According to the World Bank group [26], Modern Standard Classical Arabic is used while writing books, contracts, real articles, such as poets [26], and also in the holy Quran. The vernaculars are so hard to be translated because of their dependency on the huge experience in the Arabic Language Sciences and their interference with some other vernaculars and foreign languages. Literal translation, which means translation word by word, is not the best choice, especially when the words meanings alone cannot give us the needed meaning from the sentence at all [5,26]. Newmark stated that [2] a text may need to be contemplated before being translated in ten different directions, which can be summarized as follow: The idiolect of the author of the Source Language (SL). The regular linguistic and lexical usage for this type of text depends on the context, and content items and referring specifically to SL, or third language cultures (i.e., not SL or the Target Language (TL)). The standard structures of a text in books or newspaper are affected by tradition at the time. The expectations of the putative readership, bearing in mind their evaluated information of the topic and the style of dialect they utilize, communicated as far as the largest common factor since one should not translate down (or up) to the readership. The same of the last three points are for TL. What is being described or announced, found out or verified (the referential truth), where possible independently of the SL text and the expectations of the readership. The perspectives and biases of the translator, which may be personal and subjective, or may be social and cultural, involving the translator's 'group loyalty factor,' which may reflect the interpreter’s national, political, ethnic, religious, social class, sex, etc. assumptions [2]. Dickins et al. [5], has divided the texts to be translated into five matrices: First is Genre: this is related to the literary, philosophy, thoughts and religion. The second one is Cultural: this is related to idioms, cultural transplantations stories, and proverbs. The third matrix is Semantic matrix: this is related to meaning of synonyms, collocations, and original metaphors. After that Formal: this is related to levels of graphics, sentences, grammar, and Quranic allusions. Last but not least is Varietal: which is related to dialects, sociolects, tones and social registers [5].

On the other hand, the Target Text (TT) that is the translation is not entirely similar to the Source Text (ST) as to its structure or content because of the limitations imposed by the semantic and the formal differences between the Source Language (SL) and the Target Language (TL).

The users of the TT evaluate it from three different perspectives: which are functional, structurally and semantically. The functional perspective deals with the text as if it is really done by the author in TT. The structure perspective evaluates the sequence and the arrangement of the ideas in the TT according to the Source Text (ST). The semantic perspective is related to testing the meaning of the TT in comparison with the ST [27]. Newmark stated that [2], translation is divided according to the needed translation function into an expressive function, an informative function, a vocative function, an artistic function, a phatic function, and a metalingual function. The expressive function depends on the mind of the author and expresses, in many cases, his or her feelings. The types of these texts are Serious imaginative literature (lyrical poetry, short stories, novels, or plays), Authoritative statements (political speeches, and documents by ministers or party leaders, statutes and legal documents, scientific papers, philosophical and academic works written by acknowledged authorities), and finally, autobiography, essays, and personal correspondence. The informative function includes reported ideas or theories, and their texts are often standard. Examples of the informative texts are a textbook, a technical report, an article in a newspaper or a periodical journal, a scientific paper, a thesis, and minutes or agenda of a meeting. The following four points on a scale of language varieties describe it further: a formal non-emotive technical style for academic papers, a neutral or informal style with defined technical terms for textbooks, an informal warm style for popular science or art books, and a familiar, racy, non-technical style for popular journalism. The vocative function of the language is the audience, the addressee. All vocative texts are the relationship between the writer and the audience, and these texts must be written in a language that is instantly comprehensible to the public. These three types may be found together but rarely or binary. The artistic function is a language designed to please the senses, through its actual or imagined sounds, and through its metaphors. The metaphor is a mixture of the expressive and the artistic function. The phatic function is used for maintaining friendly contact with the receiver rather than for reporting foreign information. The metalingual function indicates the language ability to explain, label, and criticize its features [2].

Taylor stated that [17], many of the challenges while the translation was the great diversity of structures and the process of interpreting the sentence that used a specific structure and then choosing the best structure of the sentence in the target language [17]. The translation was a commercial activity in 1990 [28], and one of the first Statistical Machine Translations was developed by IBM [18] which was intended to automate the core translation task [17]. There are many types of the translations that are using the technology to aid translation, which will be covered in the following section.

## 3. MACHINE-TRANSLATION

Machine Translation is one of the important and hard to implement fields in our daily life. Bar-Hilal stated that Machine Translation (MT) had become a multimillion dollar affair [29]. It is because it needs very complex processes to achieve reasonable results. There are many types of research about building or developing Machine Translation for the English language and many other natural languages. In contrast, similar types of research for the Arabic language is very few and somehow nonexistent. For instances, the previous research papers for the Arabic Language (with other languages like Hebrew) was a Machine-Aided Human Translation (MAHT) or just a Natural Language Processing (NLP) pieces of research to develop a parser which has many translation errors. Recent Pieces of research develop a Neural Machine Translation (NMT) based on Statistical Machine Translation (SMT) whose performance is comparable to the performance of SMT, which is not satisfied anymore because of the huge number of errors found in the translated texts. Semantic web provides us with a mean to develop MTs whose performance can be satisfied compared with the performance of SMTs, according to the results provided by the researchers who developed the MT based on semantic web technologies.

Hutchins [14] stated that the MT system might be designed to satisfy the following criteria: to deal with single words, to get restricted input text structure, and to have pre-edited input texts with any grammatical set, and without caring about the ambiguity of words or any other operators during translation. Ambiguity could be either nouns, verbs, or adjectives [14], but this depends on the properties of the language itself. Hutchins also classified the MT to many groups according to the aiding type (machine-aided MT, human-aided MT, computer-aided translation). Input or output edition (pre-edited or post-edited), number of targeted languages (bilingual or multilingual), translation approach (direct translation, Interlingua, or transfer) and syntactic structure analysis (predictive analysis, phrase structure or dependency grammar). They also pointed to the importance of semantic MT, but it was just a survey, and there were no implementations for their vision [14]. Chan et al. [30], integrated a state-of-the-art Word Sense Disambiguation system into a state-of-the-art hierarchical phrase-based MT, "Hiero" [30]. In contrast, they demonstrate only one way for the integration without introducing any rules that compete against existing standards [31]. Gupta et al. [31] introduced 16 features that were extracted from the input sentence and their translation. Then showed a quality score based on Bayesian inference produced from their sample training data, but they did not develop a new English-Hindi MT, and their work was just an analysis study for an existing MT [31].

## 3.1 Statistical Machine Translation (SMT)

Statistical machine translation (SMT) is a term that refers to a group of MT systems that are developed using the machine learning methods. It is the most extensively studied type of MTs according to Srivastava et al who depicts a typical model for the SMT workflow [32]. The first SMT systems were developed less than three decades ago [33]. SMT is still the core of the new MT systems either those that are SMT in themselves, NMTs or MTs which are based on the semantic web. IBM in 1990 used Bayes theorem to develop a statistical approach to Machine Translation [18]. According to Lopez et al. [33] SMT task is to take a sequence of words, tokens or symbols in the source language with vocabulary group can be called A and transform it into a sequence of words, tokens or symbols in the target language with another vocabulary group can be called B [33]. Lopez also in [33] classified the SMT, as shown in Fig. 1, according to the translation equivalence to:

- **Finite-State Transducer Models:** extend two common finite-state automata sets (FSAs) of labels (the input and the output sets) and composed by making the output of one input to the next. Therefore, it needs to record the sentence before sharing the final translation [33]. It is also widely used in speech recognition. It was used as an efficient machine in the sequential string-to-string and the sequential string-to-weight transducers. The minimization and determinization algorithms for string-to-weight transducers were used to solve some of the theoretical issues against the machine performance [34]. This type of models has two other different kinds [33]:
  - **Word-Based Models:** translates the source language word by word or more in the target language. When the user wants to translate a sentence, the most used meaning of a word will be used to translate that word always in the sentences [33]. It can also be used as a word segmentation as it was proposed by Sun with the Chinese language [35]. It also can be used to align words that can be utilized in the phrase-based models while Machine Translation statistically [36].
  - **Phrase-Based Models:** translates the source language phrase by just a phrase in the target language. When the user wants to translate a sentence, that translation of the words will be shown in a sentence in the target language [33]. Callison-Burch et al. [37], has rescaled this model to make it correspond with the longer phrases and the larger corpora. It is a way to solve the idioms problem. Besides, it is a widely used model because the error rate is small and it is fast in comparison with all other machine translations [37]. However, this statistical or learning method is not applicable to identify the best meaning semantically because it replaces a phrase by another phrase in a different language. If there's a phrase that has a different meaning, the replacement mechanism will choose the most frequently used translation every time.

- **Synchronous Context-Free Grammar Models:** a generalization of the context-free grammar (CFG). It can be applied with syntactic parsing of one or both languages in translation [33]. It can be used with the phrase-based model to develop the hierarchal phrase-based model which is used while building the Neural Machine Translation (NMT) [38]. It also can be used with the tree transducers promise to enhance the quality of the output of the SMTs where the complexity is exponential due to arbitrary reorderings between the two languages [39]. Decoding can be more flexible to synchronous context-free grammars by enhancing the unary rules by binarizing rules and phrases [40]. This type of models has many other different types [33]:

- **Bracketing Grammars:** strict the number of rearrangements described by the synchronous context-free grammar model. However, the model sometimes cannot express the reordering relationship [33]. It is also used to induce the structure of the analogous translation grammar in the German-English alignment. It is done to disambiguate the meaning and to correct the bracketing errors [41,42].
- **Syntax-Based Grammars:** this is used in noisy-channel machine translation [43] to easily merge the knowledge based on the natural language grammar rules. The needed result is to reorder the observations of the phrasal cohesion constraint imposed by exact linguistic syntax [33].
- **Hierarchical Phrase-Based Translation**: like word-based models, it only allows word-to-word translation, then requires a reordering decision for each word [33], which may cause errors for the languages that add some letters to some words to complete the meaning of the sentence. Each decision leads to more possibility for error. Explicit word alignments are not specified in this model, although it implies that there may be some an internal alignment between words appearing in the same grammar rule [33]. It is formally a synchronous context-free grammar but is learned from a parallel text without any syntactic annotations [38].
- **Syntactic Phrase-Based Models:** involve multi-word translation rules, as in phrase-based translation models. These are decorated with fragments of linguistic syntax which are used to constrain the possible reordering of the phrases [33]. Using a large n-gram language models is ideally the essential part in the best-performance of the Machine Translation Systems [44]. Incrementing the syntactic parsing into phrase-based translation will re-make the role of the language model as a mechanism for encouraging syntactically fluent translations [45].
- **Alternative Linguistic Models (synchronous dependency grammar)**: considers the nodes of the rooted tree, those are describing the sentence structure, words of the sentence. It is possible to derive conversions that will convert many dependency grammars to context-free grammars and vice versa [33]. Dependency structures are naturally lexicalized as each node is one word. In contrast with phrasal structures (treebank style trees) which have two node types: lexical item terminals, and nonterminals that store word orders and phrasal scopes. A synchronous derivation process cannot deal with two types of cross-language mappings: crossing-dependencies and broken dependencies. A synchronous derivation process for the two syntactic structures of both languages proposes the level of cross-lingual isomorphism between the two trees of those structures [46].

❖ **Other Models of Translational Equivalence (tree-adjoining grammar):** one of a large class of formalisms known as linear context-free rewriting systems (LCFRS). These formalisms can parse a constrained subset of context-sensitive languages in polynomial time. Much of the research in this area is still theoretical [33].

The statistical method was introduced as a new approach for MT in 1949 [47]. Statistical MT was stated by IBM researchers in 1986 [18]. They thought that MT is as old as the first generation of the computers. They also said that the translation must depend on many factors and the essential one was the whole original text itself. In contrast, they treated the words without recognizing the connection between words or even recognizing the sentence structure. Koehn et al. [13], stated that learning phrases longer than three words and learning phrases from high-accuracy word-level alignment models do not have a strong impact on performance. However, learning only syntactically motivated phrases degrades the performance of their systems.

## 3.2 Machine Translation using Semantic Web (SWMT)

A semantic web restructures the massive amount of data that is accessible and understandable to both humans and machines on the internet in a way that is similar to that of the human mind. It would be like training the internet to understand the context that is close to whatever word or phrase being searched through tags, the searcher attaches to the subject. The semantic web would serve as a way that helps computers think like the way humans do while still allowing the humans to understand those same files [48]. Using the semantic web, machines have the ability to deduce new facts from existing ones according to the available data. Besides, the semantic web enables computers to store and retrieve information and also produces new information. The ontology is the way that computers can understand the information they are carrying [48].

Moussallem et al. [8], as shown in Fig. 2, classifies the MT according to their architecture, methodology, problem space addressed, and performance to:

- **Rule-Based Machine Translation (RBMT):** parsing the input, creating an intermediary linguistic representation, generating a text in the target language based on the morphological, syntactic and semantic mapping between the two languages.
  - **Direct approaches:** in which words are being translated one by one, without considering the variations in meaning of the words. This leads to significant error rates.
  - **Transfer-based approaches:** analyze the sentence structure of the source language, transfer that sentence to an intermediary structure, and then generate an internal representation based on linguistic rules of the target language. Thus, it uses three dictionaries: two monolingual dictionaries and a bilingual dictionary from the source to the target.
  - **Interlingua-based approaches:** extracting the language independent representation of the input text and achieve a higher accuracy than comparable linguistic methods.
- **Corpus-Based Machine Translation:** commonly use a large dataset containing the previous translations or a large set of examples to tackle the translation task. A bilingual corpus is essential for it.
  - **Statistical approaches:** this model is based on this type of MT. This methodology uses machine learning techniques including supervised, unsupervised, and semi-supervised approaches to compute a statistical translation model from input corpus and use it in subsequent translations. SMT requires large bilingual corpora to achieve good accuracy. Hence, the translation quality directly depends on the reference corpus selected for the training.
  - **Example-Based approaches (EBMT):** uses bilingual corpora, but they store different data. The training phase imitates a basic memory technique of human translators and is akin to filling the database of a translation memory system. EBMT task consists of source text decomposition, translation examples matching and selection, and adaptation and recombination of the target translation [49].
- **Hybrid Machine Translation:** a group of both the rule-based and the corpus-based MTs.

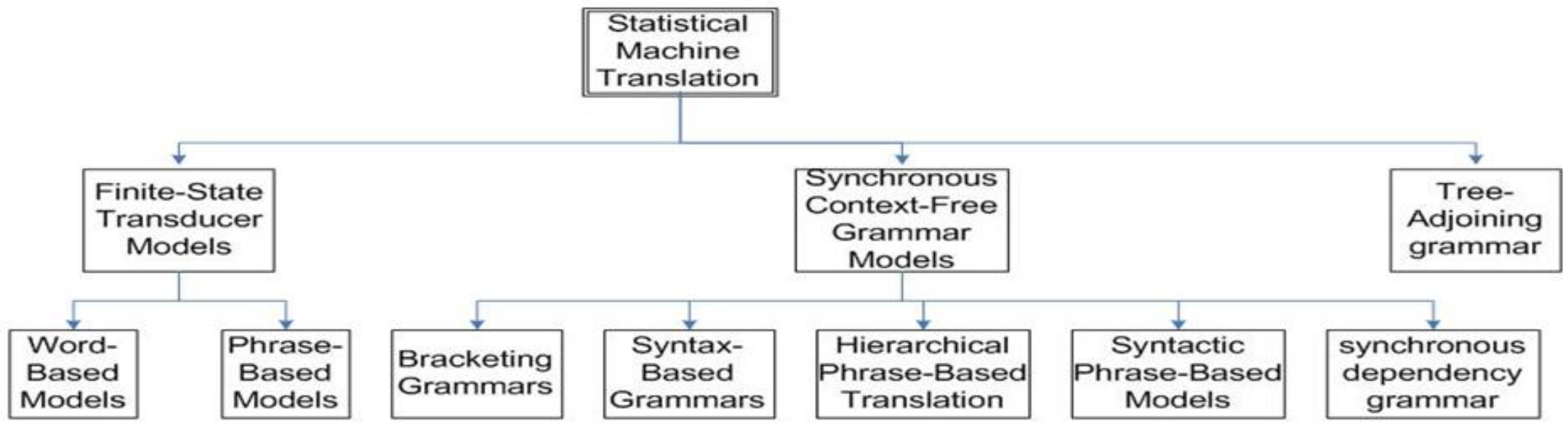

**Fig. 1. Statistical machine translations classification**

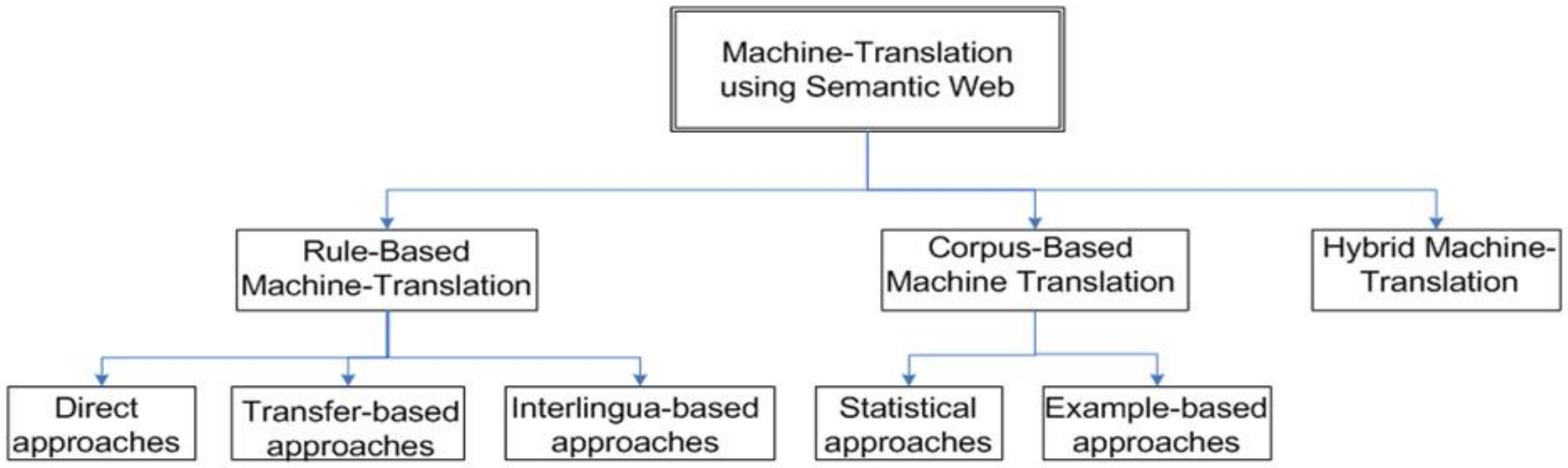

**Fig. 2. Classification of machine translations using semantic web**

Shi et al. developed a world model for Chinese-English MT using an Ontology-driven [50]. Seo, et al. [19], presented a syntactic and semantic method for English-Korean MT using Ontology for web-based MT [19]. Mousallem et al. [12], solved the ambiguity in dialog conversation using an ontology-based MT [12]. However, all of them can only serve for their proposed languages. They also mentioned that the instant translators such as google, Bing, and more others are statistical translators and have no semantic processing in their works [12]. Alagha et al. [51], presented a domain-independent approach to translating Arabic NL queries to SPARQL by getting benefits from the linguistic analysis [51]. It was just to translate the Arabic questions to SPARQL, not to other human-languages. The systems developed in [52,53] taking questions expressed in Arabic and returns the answers drawn from an ontology-based knowledge base through the ontology file was not prepared as the MT system needs. While the translation is done to the SPARQL level only, it would not work as an MT between human-languages, and it would be like the previous works that didn't translate the Arabic question sentences to any other human-languages.

## 3.3 Neural Machine Translation (NMT)

Neural MT (NMT) "is a recently active topic that conducts the automatic translation workflow very differently with the traditional phrase-based SMT methods. Instead of training the different MT components independently, NMT model uses the Artificial Neural Network (ANN) to learn the model jointly to maximize the translation performance through two steps recurrent neural network (RNN) of both the encoder and decoder" [54-57]. Dowmunts et al. [58] made a comparison between the phrase based machine translation and the neural machine translation and found that they are equal, or the neural one gets somehow higher results than the phrase based. This is because the neural machine translation is used to remove the ambiguity using the n-gram technique and the results are to be sent to the phrase based machine translation to complete the translation. Besides, NMTs crucial goal is to build a single Neural Network that can be mutually adjusted to increase the translation performance [55].

To the best of our knowledge, the Neural Machine Translation (NMT) is an encoder-decoder architecture, according to Zhang et al. [59], proposed by Blunsom and Cho et al. in [55, 60] respectively. NMT is an end-to-end MT that consists of two RNN [59], one is the encoder and the other is the decoder. The encoder network links each source sentence with a context vector representation according to a variable length of that sentence, and the decoder network produces the targeted translation word-by-word beginning from that context vector [59]. They proposed two approaches: handling source-side monolingual corpus in SMT and exploiting target-side monolingual data in NMT. The source-side monolingual corpus is to use the source-side monolingual corpora to strengthen the encoder model of NMT, but they didn't fully explore it and they have no significant BLEU gains that are reported. Almahairi et al. [11], develop their first NMT as a neural morphological analyzer based on a phrase-based statistical machine. The performance of that machine was comparable with the performance of the phrase-based statistical machine or slightly better. Dowmunt, et al. [58] studied the neural machine translation based on the parallel corpus of six United Nations languages and phrase-based machine translation and has similar performance of Almahairi's. However, NMT cannot remedy a larger vocabulary due to the increase of the training and the decoding complexity proportionally with the number of target words [61]. Sennrich et al developed Nematus which is a toolkit that prioritizes high accuracy, usability, and extensibility of the NMT. Nematus has been used to "build top-performing submissions to shared translation tasks at WMT and IWSLT, and has been used to train systems for production environments" [62].

More examples of NMT include ByteNet and the Character-Based NMT. Kalchbrenner et al. [63], developed an NMT to run in a linear time named ByteNet as a stack of two dilated convolutional neural networks, these two are the encoder and the decoder. Because the English and German strings encodings are sequences of characters, no clear division into words or morphemes is applied to the strings. The outputs are strings of characters in the target language [63]. Costa-juss`a et al. [64], proposes a character-based neural MT system which uses the character transformation representations which is embedded in a mixture of convolutional and highway layers. It is to increase the performance in many NLP tasks, including machine translation [64]. OpenNMT is another example of NMT implementation that prioritizes efficiency and modularity. [65].

Machine Translation using the deep learning (DLMT) seems like a branch of the NMT, but it differs from the NMT which is an end-to-end machine learning approach [66], by depending on pattern recognition, too [67]. In contrast, Most of the existing NMT models are shallow, and this makes a performance gap between the single NMT model and the best conventional MT system [68]. Also, NMT systems are computationally expensive both in training and in translation inference. Therefore, Google uses the DLMT to bridge the gap between the human translation and the machine translation by developing a model that uses deep Long Short-Term Memory (LSTM) networks with beside eight encoder and eight decoder layers via residual connections as well as attention connections from the decoder network to the encoder ones. They use an attention mechanism to connect the bottom layer of the decoder to the top layer of the encoder to improve parallelism, and as a result, to decrease training time. They employ low-precision arithmetic through the inference computations to accelerate the final translation speed. Rare words were handled by dividing the words into a limited set of common sub-word units ("word pieces") for the input and the output texts [66]. However, the gap was just reduced but not completely avoided.

As a result of all those types of research, researchers are still trying to find a way that enables the machine to provide us an accurate translation. We summarized the previous studies of the above three types of MT in Table 1 which is mentioned below. Because the SMT depends on pre-stored words or phrases translation [12, 33] and the NMT depends on the n-gram mechanism [11, 58], their translations sometimes are better than SWMT [4]. However, they are all bad in metaphor translation because metaphor needs to be looked at from many sides [4] before providing the final translation. A comparison between the translation performance quality of the SMT, SWMT, and NMT shown in Table 2.

**Table 1. MT types summarization**

| | SMT | SWMT | NMT |
|---|---|---|---|
| **Based on** | Statistical rules and Machine Learning (ML) Algorithms, eg: EM, Naïve, or Cluster algorithms. It needs a syntax tool that's work differs according to the type of SMT. | Semantic analysis using semantic tools with or without a syntax analysis tools. | Statistics and Neural Network (NN) or Deep Learning (DL) algorithms, which is a type of ML. It differs from SMT by using encoder and decoder parts, beside SMT part, to work as syntax analysis tools. |
| **Mechanism** | Take a sequence of words, tokens or symbols in the SL with vocabulary group A and transform it into a sequence of words, tokens or symbols in the TL with | Texts are semantically then syntactically analyzed or vice versa. If text aren't analyzed syntactically, then it will have a morphological analysis. Both syntactical | Using an encoder RNN to encode a sequence of characters into words or morphemes then send it to a SMT to look for the equivalent words or |

| | SMT | SWMT | NMT |
|---|---|---|---|
| | vocabulary group B | and morphological analysis have a dictionary file to facilitate the translation process. Semantic side uses annotation, SPARQL, or reasoner method. | morphemes then send them to the decoder RNN to decode the characters in the TL. |
| **Emergence** | First introduced in 1949 as a new approach and the most rapidly growing one. Almost all the other types depend on it or on its techniques. | Somehow in the same time of the SWT. However, because the ontology was first built in 1990s where it is a core component, SWMT seems like it has emerged in late time. | In the most recent few years. We use recent research those are from 2013 and above. |
| **Accuracy** | If the translated texts were recorded before, the accuracy will be 100%, if no, the accuracy will be noticeably bad. | When the semantic side is used to correctly choose the right meanings of the homograph, the accuracy will be noticeably better than SMT. | Equal to or comparable with SMT. |
| **Special feature** | Saves SL and TL as an equivalent pair of texts. | Doesn't need to save everything just saves only the semantic meanings. | remove the ambiguity using the n-gram technique |

**Table 2. Translation performance quality of the three types of MT**

| Phenomenon | SMT | SWMT | NMT |
|---|---|---|---|
| Homograph | Bad | Excellent | Bad |
| Terms | Good | Somehow bad | Somehow good |
| Idioms | Excellent | Bad | Excellent |
| Phrasal verbs | Very bad | Very good | Bad |
| Imperatives | Somehow good | Good | Good |

## 4. SOME NEEDED TOOLS

This section is going to show the different research papers that cover the components we are going to use in the proposed model. We are going to talk about the Arabic parsers, the morphological analyzers, the semantic analyzers and the correction algorithms. The parsers may contain some necessary tools such as the context identification and disambiguation, and the Named Entity Recognizers (NER). The semantic analyzers depend on the Ontology files.

### 4.1 Arabic Parsers

As mentioned before, the Arabic language has various structures. According to Ba-Alwi et al. [10], the Arabic words combine two morphemes: the root which is a set of radicals and the pattern which is a set of vowels. These vowels with the radicals form the word stem [10]. Besides, the words cannot always form the sentences in their stems' forms only, but also there may be more additional letters those may give another meaning to the word. Natural Language parser is a machine that can understand the sentence parts and serves us in the translation while using the MT. Many researchers have worked on this type of machines. In contrast, in this research, we are going to explore some of the most modern Arabic Parsers which make a real differentiation in Arabic translation and Arabic Natural Language Processing.

Arabic Parser was proposed as an NLP tool that is used to analyze the sentence into its structural components [20,69-71]. The English language has just one structure to construct the sentence, i.e. subject (S) then verb (V) and after that is the sentence complement or the object (C, O). However, Al-Raheb et al. [72], stated that Arabic is a free word order: SV(C, O), VS(C, O), V(C, O), S [72]. In our vision, the structure of the Arabic sentence can be different from what they have stated. Also, according to Green et al. [73], their parser was similar to another Treebank in gross statistical terms; annotation consistency remains problematic [73]. Tounsi et al. [74], thought that the best-known Arabic statistical parser, at that time, was developed by Bikel in [75]. Thus they tried to enrich that parser's output with more abstract and profound dependency information. On the other hand, the categories they added to Bikel's parser resulted in substantial data-sparseness [74]. Zaghouani et al. [76], added new specified post-editing rules of correction, according to seven categories: spelling errors, word choice errors, morphology errors, syntactic errors, proper name errors, dialectal usage errors, and punctuation errors [76].

A simulation example of the parsers is Microsoft Arabic Toolkit Service (ATKS) as shown in Fig. 3. Microsoft developed another toolkit for language analysis called MSR SPLAT. It includes both traditional linguistic analysis tools such as part-of-speech taggers and more recent developments such as sentiment detection and linguistically valid morphology [77]. The available functionalities in this toolkit are part-of-speech taggers, chunkers, parsers both, the Constituency and the Dependency, Semantic Role Labeling, Sentence Boundary / Tokenization and Stemming / Lemmatization, and they provide it as a web service English parser.

According to Microsoft Arabic NLP toolkit developer, ATKS has a similar technology of that used in MSR SPLAT. However, there are no published papers that cover ATKS parser. ATKS parser was employed by Al-agha et al in [51-53] to simplify the process of converting the Arabic sentence into the RDF triples. Al-agha et al. [51], stated that ATKS has many functionalities over than others [51]. However, it may produce errors while parsing. The correct parsing of the sentence is shown in Fig. 4. According to Fig. 3, the sentence "عاد أخي الأكبر عمر ومعه صديقه" has two sentences connected with the preposition "و", which leads to an error because the first one has no complement and the second one is a complement with no subject or object, where the two parts besides the preposition "و" are only one sentence. Green et al. [73], introduced the Stanford gold parser that can be used to parse any sentence in one of the united nation formal six languages (Arabic, English, French, Spanish, Chinese, Russian). Their proposed parser, according to Stanford University, is demonstrated as one of the most efficient and found parsers [73]. Stanford parser can distinguish the POS tags of the English sentences successfully with high precision. By an experiment, while parsing six paragraphs only one sentence was badly assigned to incorrect POS tags because the words in that sentence are commonly used as POS tags rather than those used in the sentence we parsed. For the Arabic Language, they also showed that, in their paper, Arabic parsers are poorer in thoughts and still much lower than English ones [73]. Using Stanford parser [1] to parse the same sentence that we used to test ATKS, it can distinguish the two parts as being one sentence, as shown in Fig. 5, if we identify the parts of each word before sending the sentence to be parsed. But, unfortunately, it cannot segment the words into its really component by its own. Farasa parser[2] is another parser example that can be used to parse an Arabic sentence. Although it can identify some POS tags successfully, it didn't generate a parse tree which will lead to a problem while translation because the words in the sentence will be all at the same level. Xerox parser[34] can also give us very important results. It is an ambiguity discovering tool rather than just being a parser. It reads the sentence words and provide us with every possible POS tags of the single word. In contrast other parsers, it neither generates a parse tree nor gives us the POS tags of the words according to their position in the given sentence. Stanford CoreNLP parser has ideal results while parsing English sentences. However, Arabic sentences still have no parser that can do parsing with correct results such that was done with the English sentences. A comparison between the different parsers and their functionalities with the Arabic sentences is found in Table 3.

#### 4.1.1 Context identification and disambiguation

Parsers also can be used to define the contexts. According to the best of our knowledge, the most important component of any Machine Translation is the context identification mechanism, which is an effective way for disambiguation [12]. Abdolahi et al. in [78] think that contexts could be specified using the local coherent sentences. They also mentioned that Soricut et al. [79], proposed to use EM algorithm to determine the local coherence because EM can be used to choose the most likely synonym or equivalent to each word. Thus, the keywords are the connections that relate some words in many other sentences. Alrakaf et al. [80], stated that the word with its different diacritics is the main reason for the ambiguity in the sentence which affects the accuracy, efficiency, and effectiveness in the Machine Translation. Finally, they used the Naïve Bayes Classifier to find the most accurate meaning of the ambiguous words.

[1]http://nlp.stanford.edu:8080/parser/index.jsp
[2]http://qatsdemo.cloudapp.net/farasa/demo.html
[3] Arabic: https://open.xerox.com/Services/arabic-morphology/Consume/Morphological%20Analysis-218
[4] English: https://open.xerox.com/Services/fst-nlp-tools/Consume/Morphological%20Analysis-176

Zhonglin et al. [81], stated that Part-of-speech Tagging is the core of the NLP. The traditional statistical machine learning methods of POS tagging depends on the high quality of training data. However, it is very time-consuming to obtain that data. The methods of POS tagging, based on dictionaries, ignore the context information which leads to lower performance [81]. They provided some algorithms to modify the dictionary and to develop the POS Tagging ways. It is to develop the ways of context detection. To avoid the bad translation for the homonyms words they use two tagging ways to discover the ambiguity in the sentence to solve it. The first is using the dictionary to tag each word in the sentence, and the other is using the Maximum Entropy (ME) model to re-tag those words again. The ME model is built in the Stanford Parser and is used to choose the best tag. Once they have two different tags for the same word, then that is an ambiguous word. Thus the context information must be taken into account while translating those words. The other words translations will be taken from the dictionary directly [81]. Also, they compared their work with the other POS tagging methods available at that time and stated that their method is better and its accuracy is 88%. However, it was developed in Chinese-English dictionaries only.

#### 4.1.2 Named entity recognition

Another part of this context is the Named Entity Recognition. This term is used to assign each word in a document into predefined name entity classes. "NER classifies the proper nouns into different classes such as names of the person, place, organization, numbers, quantities, and time expressions" [82]. Essentially, NER has three types of approaches: Rule-Based or Handcrafted Approach which uses gazetteer list of triggered words. Automated or Statistical Approach which uses a list of models and algorithms like: Hidden Markov Model (HMM), Maximum Entropy Models, Decision Trees, Support Vector Machines (SVM), Conditional Random Fields (CRF), etc., and the Hybrid Approach which is a combination of the rule-based approach and one or more models or algorithms of the Statistical approach. Some of the difficult problems of NER are the language free-order, inflectional and morphological nature of the language. Besides, some languages have no capitalization to distinguish the proper names as the English way. These problems are found in some languages like the Indian [82] and the Arabic language.

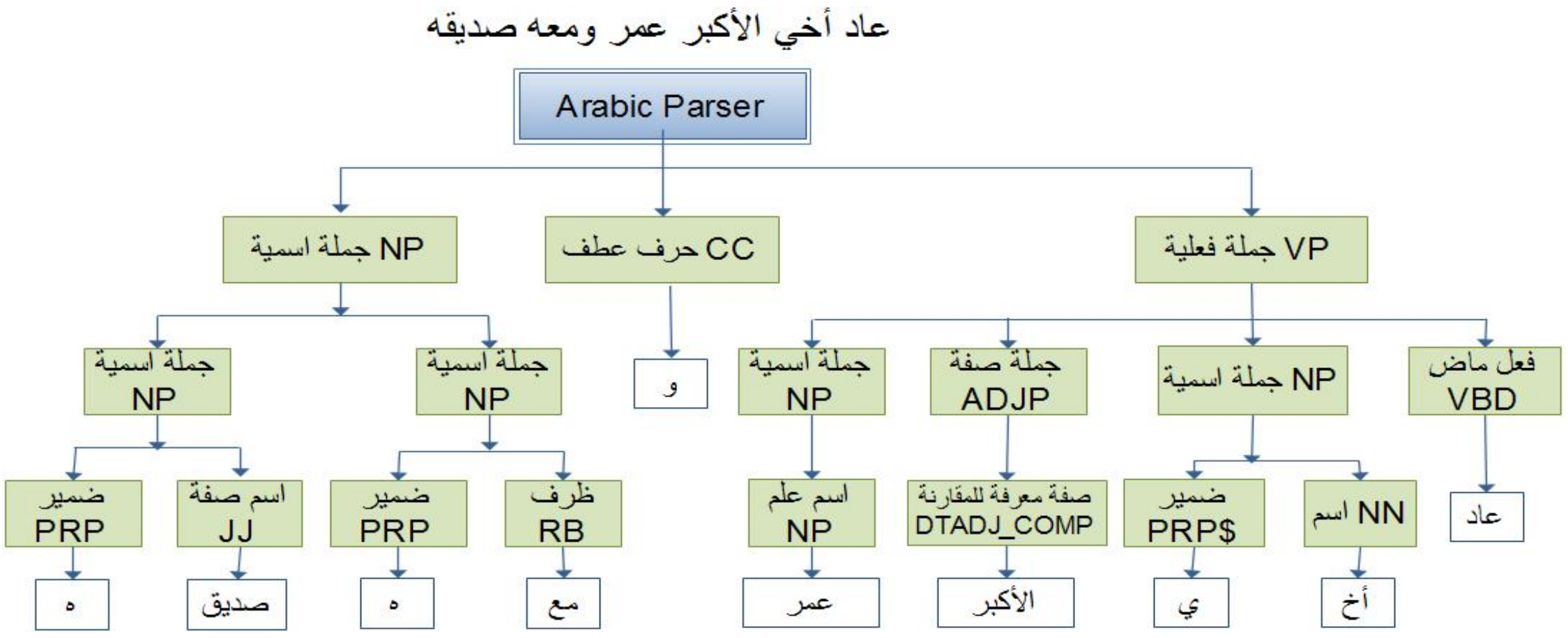

**Fig. 3. Microsoft's ATKS Parser analysis for Arabic sentence**

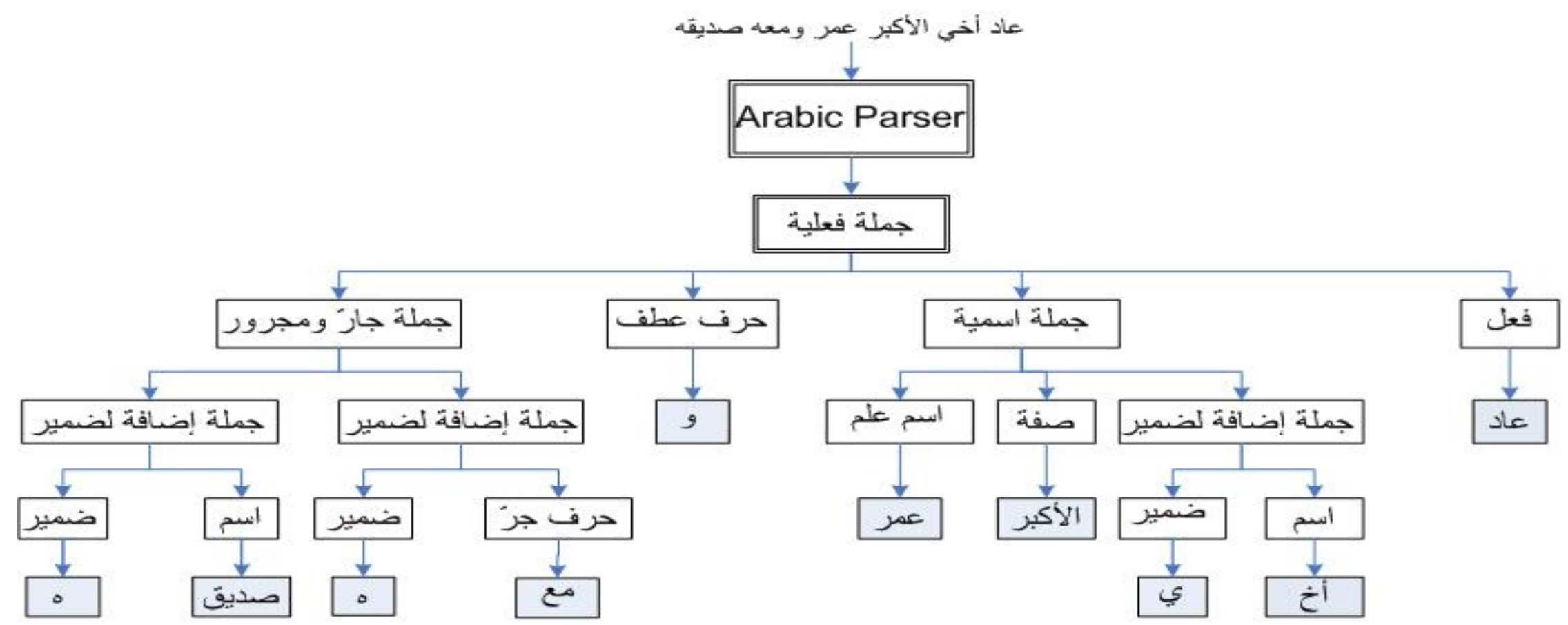

**Fig. 4. Parsing example**

Parse

```
(ROOT
  (S
    (VP (VBD عاد)
      (NP
        (NP (NN اخي)
          (NP (DTNNP الاكبر) (NNP عمر)))
        (CC و)
        (NP (NN مع)
          (NP (PRP ه))))
      (NP (NN صديق)
        (NP (PRP$ ه))))
    (PUNC .)))
```

**Fig. 5. Stanford parsing example**

**Table 3. Arabic parsers**

| Parsing process | ATKS | Stanford CoreNLP | Farasa | Xerox |
|---|---|---|---|---|
| Parsing precision | High | Low | Low | give all possibilities |
| Working with no help | Yes | No | Sometimes | Yes |
| Parser tree errors | Has errors | Very small errors | No parse tree | No parse tree |

## 4.2 Morphological Analyzer

Sawalha et al. [23] stated that 'Morphology' is the study, identification, analysis, and description of the minimal meaning bearing units (morpheme) that constitute a word. Morphological ambiguity is a major concern for syntactic parsers, POS taggers, and other NLP tools [24]. The morphological analysis gives the most relevant information for a part-of-speech tagger to select the most suitable analysis for a given context [23]. According to [22], it is a component of the parsers. However, [21] stated that a parser is a morphological analyzer tool. According to Moussallem et al. [12], English and Portuguese languages must have the same structure of the sentence parts and, as a result, there was no need to have two different parsers to process the natural language. In that research, the morphological analysis was sufficient to do the analysis, in which the sentence will be divided into its main parts, and they stated here that the homographs might be either a noun or a verb. Mousallem stated that they used Stanford CoreNLP parser as a morphological analyzer [12]. Manning et al. [21] developed Stanford CoreNLP parser by improving their last parser and providing some additional features. They stated that it is better than their past and the other available parsers at the development time.

## 4.3 Semantic Analyzer

The essential issue that MT systems have to handle is the ambiguity [8]. Therefore, from a formal point of view, this means that the goal of MT is to transfer the semantics of text from an input language to an output language [16]. Therefore, to have expert systems, knowledge needs to be recorded, processed, stored, retrieved, reused, and communicated. Ontologies are the way to do all that, and from this point, the term semantic has come from [48]. “Ontology is an explicit specification of conceptualization” [83]. “It defines the terms with specified relationships between them and can be interpreted by both humans and computers” [84]. Al-Muhtaseb [85] introduced some differences between the Arabic and English languages to propose from these differences an upper model of the Arabic Ontology found that time. The way he suggested to do that is the enhancing of the available ones to include the Arabic.

### 4.3.1 Ontology

Bella et al. [86] stated that ontology development approach is either domain-agnostic or are developed for a specific domain [86]. Al-Muhtaseb [85] developed an upper ontology model to simplify the interface between domain-specific knowledge and general linguistic resources while providing a domain- and task-independent classification system that supports natural language processing [85]. Zaghouani et al. [87], stated that punctuations make the languages and their semantics vary. Therefore they developed a punctuated Arabic corpus to be used in Arabic-English translations. According to Al-agha [51], the Arabic ontology is not fully entirely Arabic corpus. However, each stem or word in this ontology can have multi-labels or multi descriptions those we can use to add the Arabic translation to each stem and can be used in bilingual processing. Rafalovitch et al. [88], described a six-ways parallel public-domain corpus from which many of the available ontology has started. This six-ways parallel public-domain corpus contains vocabularies of six languages: Arabic, Chinese, English, French, Russian, and Spanish. This corpus is richened every moment and has been used in many types of research in the last few years.

Ontology as a type of knowledge bases needs to be inferred and its concepts need to be matched [83]. Ontology matching process includes the computations of labels and nodes formulas, then doing the matching using the background knowledge axioms. Then matching the two trees of labels and nodes [86]. Knowing the steps of matching or inference is important to validate and evaluate the generated ontology file. Gurk et al. [89], stated that detecting the problems and building the ontology in the right way reduces the costs and efforts of maintenance. In their perspective, quality assessment metrics of any ontology are the accuracy, completeness, consistency, currentness, understandability, and the availability. They also specify the accuracy dimensions which are an incorrect relationship, merging of different concepts in same class, hierarchy overspecialization, using a miscellaneous class, a chain of inheritance, class precision, and a number of deprecated classes and properties. The completeness dimensions are the number of isolated elements, the missing domain or range in properties, the class coverage and a relation coverage. The consistency dimensions are the metric dimension reference, the number of polysemous elements, including cycles in a class hierarchy, missing disjointness, defining multiple domains/ranges, creating a property chain with one property, lonely disjoints, tangledness (two methods), and the semantically Identical classes. The currentness dimension is the freshness. The compliance dimension is no OWL ontology declaration, the ambiguous namespace, the namespace hijacking, and the number of syntax errors. The understandability dimensions are missing annotations, property clumps and using different naming conventions. The availability dimensions is the dereference ability [89].

## 4.4 Sentence Reorder

MT must make sure of the precision of the translation of the target language, either from English to Arabic or from Arabic to English. Therefore, in both cases, the sentence structure must be checked. In [12], the Portuguese and English sentences can be both dealt with as if they have only the same structure. Also, it was not necessary to the sentence reordering before the final generating of the target message because the sentence will be already ordered after the statistical translation. According to that, the automatic replacement will never generate any error. In contrast, the Arabic language has different structures. The automatic replacement will replace the homonyms in their places with their relative translation according to the context. Whenever the statistical MT translates the sentence, the sentence order will be wrong. Reordering is an essential task in SMT that significantly has an impact on the quality of the generated translation [90]. There are types of research those are interested in the sentence correction, such as that proposed by Simha et al. [91], and which can be used while the text summarization and translation to make sure that the sentence is following a valid logical structure in the target language.

## 5. CONCLUSION

Translation is the main activity in our daily lives because of its importance in traditional activities, studying, and friends' conversation, and sometimes in political, and legal contracts, too. Translation has many aspects, types, and features. Translation has been recently a commercial activity, especially with the rapid acceleration in ICT which requires developing machines and tools those facilitate the communication process between the both sides. Machine Translation (MT) was one of those needed machines if it was not the hidden soldier in any communication process. However, MT could not until this moment transcend the human translation. However, fortunately, the research has not stopped in this area and hopefully will continue to contribute in this field. There are three types of MT: SMT, SWMT, and NMT. Sometimes, some other tools are needed to be used during MT, to help analyze the text and then develop the texts in the target languages. These types of research are continuing and will never stop till the MT could be equal to or better than the human ones.

As a future work, hybrid research of the three types may be needed to be developed to benefit the advantages of all the MT types discussed in this research. The English language was the main language in many bilingual or multilingual MT groups of research. New studies can be done for many other languages except the English and without using it as a pivot language. New studies may be developed because the needed tools and toolkits are also still hot topics.

## COMPETING INTERESTS

Authors have declared that no competing interests exist.

## REFERENCES

1. Luo Y, Bu J. How valuable is information and communication technology? A study of emerging economy enterprises. Journal of World Business. 2016;51:200–11.
2. Newmark P. A textbook of translation. New York: Prentice hall; 1988.
3. Jakobson R. On linguistic aspects of translation. On Translation. 1959;3:30-9.
4. Macketanz V, Avramidis E, Burchardt A, Helcl J, Srivastava A. Machine translation: Phrase-based, rule-based and neural approaches with linguistic evaluation. Cybernetics and Information Technologies. 2017;17(2):16.
5. Dickins J, Hervey S, Higgins I. Thinking Arabic translation: A course in translation method: Arabic to English. Cambridge, UK: Routledge; 2002.
6. Hatim B, Munday J. Translation: An advanced resource book: Psychology Press; 2004.
7. Martin JH, Jurafsky D. Speech and language processing. International Edition. 2000;710.
8. Moussallem D, Ngomo A-CN, Wauer M. Machine Translation Using Semantic Web Technologies: A Survey. IOS Press. 2016;1(5):1-22.
9. Indurkhya N, Damerau FJ. Handbook of Natural Language Processing. Second Edition ed. R. Herbrich TG, editor. Microsoft Research Ltd.: Cambridge, UK; 2010.
10. Ba-Alwi FM, Albared M, Al-Moslmi T. Choosing the Optimal Segmentation Level for POS Tagging of the Quranic Arabic. British Journal of Applied Science & Technology. 2017;19(1):10.
11. Almahairi A, Cho K, Habash N, Courville A. First Result on Arabic Neural Machine Translation. arXiv preprint arXiv. 2016;1606(02680).
12. Moussallem D, Choren R. Using Ontology-Based Context in the Portuguese-English Translation of Homographs in Textual Dialogues. International Journal of Artificial Intelligence & Applications. 2015;6(5):17-33.

13. Koehn P, Och FJ, Marcu D, editors. Statistical phrase-based translation. Proceedings of the 2003 Conference of the North American Chapter of the Association for Computational Linguistics on Human Language Technology; 2003: Association for Computational Linguistics..
14. Hutchins J. Machine translation: History and general principles. The encyclopedia of languages and linguistics. 1994;5:2322-32.
15. Bowker L. Computer-aided translation technology: A practical introduction. University of Ottawa Press; 2002.
16. Hutchins WJ, Somers HL. An introduction to machine translation. London: Academic Press. 1992;362.
17. Taylor C. The Routledge Handbook of Translation Studies. Revised Edition ed. Munday J, editor: Routledge; 2009.
18. Brown PF, Cocke J, Pietra SAD, Pietra VJD, Jelinek F, Lafferty JD, et al. A statistical approach to machine translation. . Computational linguistics. 1990;16(2):79-85.
19. Seo E, Il-Sun S, Su-Kyung K, Ho-Jin C. Syntactic and semantic English-Korean machine translation using ontology. Advanced Communication Technology, 2009 ICACT 2009 11th International Conference on; 15-18 Feb. 2009: IEEE; 2009.
20. Othman E, Shaalan K, Rafea A, editors. A chart parser for analyzing modern standard Arabic sentence. Proceedings of the MT summit IX workshop on machine translation for Semitic languages: issues and approaches; 2003; USA.
21. Manning CD, Surdeanu M, Bauer J, Finkel JR, Bethard S, McClosky D. The Stanford CoreNLP natural language processing toolkit. ACL (System Demonstrations); 2014.
22. Zaghal R, Sbeih AH. Arabic morphological analyzer with text to voice. In Communications and Information Technology (ICCIT), 2012 International Conference on; 26-28 June 2012; Hammamet, Tunisia: IEEE. 2012;48-52.
23. Sawalha M, Atwell E, Abushariah MA. SALMA: standard Arabic language morphological analysis. Communications, Signal Processing, and their Applications (ICCSPA), 2013 1st International Conference on: IEEE; 2013.
24. Attia M. An ambiguity-controlled morphological analyzer for modern standard arabic modelling finite state networks. Challenges of Arabic for NLP/MT Conference,; London, UK. : The British Computer Society; 2006.
25. Venuti L. The translator's invisibility: A history of translation. : Routledge; 2008.
26. Group TWB. The World Bank Translation Style Guide. Arabic Edition ed. Group TWB, editor. Washington, United States of America: The International Bank for Reconstruction; 2004.
27. Kuzenko GM. The World of Interpreting and Translating. Mykolayiv: MSHU. Peter Graves; 2008.
28. Boucau F. The European Translation Markets: Updated facts and figures: Brussels: EUATC; 2006-2010.
29. Bar-Hillel Y. The present Status of Automatic Translation of Languages. Advances in Computers. 1960;1:91-163.
30. Chan YS, Ng HT, Chiang D. Word sense disambiguation improves statistical machine translation. Annual Meeting-Association for Computational Linguistics. 2007;33.
31. Gupta R, Joshi N, Mathur I. Analyzing quality of English-Hindi machine translation engine outputs using Bayesian classification. International Journal of Artificial Intelligence and Applications. 2013;4(4):165-71.
32. Srivastava A, Rehm G, Sasaki F. Improving machine translation through linked data. The Prague Bulletin of Mathematical Linguistics. 2017;108:12.
33. Lopez A. A survey of statistical machine translation (No. UMIACS-TR-2006-47). Maryland Univ College Park Dept of Computer Science; 2007.
34. Mohri M. Finite-state transducers in language and speech processing. Computational Linguistics. 1997;23(2):269-311.
35. Sun W, editor Word-based and character-based word segmentation models: Comparison and combination. Proceedings of the 23rd International Conference on Computational Linguistics: Posters 2010: Association for Computational Linguistics.
36. Lopez A, Resnik P, editors. Word-based alignment, phrase-based translation: What's the link? Proc of AMTA 2006; 2006.
37. Callison-Burch C, Bannard C, Schroeder J, editors. Scaling phrase-based statistical machine translation to larger corpora and longer phrases. Proceedings of the 43rd Annual Meeting on Association for Computational Linguistics 2005: Association for Computational Linguistics.
38. Chiang D, editor A hierarchical phrase-based model for statistical machine translation. Proceedings of the 43rd Annual Meeting on Association for Computational Linguistics; 2005: Association for Computational Linguistics.
39. Huang L, Zhang H, Gildea D, Knight K. Binarization of synchronous context-free grammars. Computational Linguistics. 2009;35(4):559-95.
40. Chung T, Fang L, Gildea D, editors. Issues concerning decoding with synchronous context-free grammar. Proceedings of the 49th Annual Meeting of the Association for Computational Linguistics: Human Language Technologies: short papers; 2011: Association for Computational Linguistics.
41. Carl M. Inducing Translation Grammars from Bracketed Alignments: Springer Netherlands; 2003. 339-61 p.
42. Zhang J, Liu S, Li M, Zhou M, Zong C, editors. Mind the Gap: Machine Translation by Minimizing the Semantic Gap in Embedding Space. Proceedings of the Twenty-Eighth AAAI Conference on Artificial Intelligence; 2014; China: Zong Publications.
43. Charniak E, Knight K, Yamada K, editors. Syntax-based Language Models for Statistical Machine Translation. Proceedings of MT Summit IX; 2003.
44. Brants T, Popat AC, Xu P, Och FJ, Dean J, editors. Large language models in machine translation. Proceedings of the Joint Conference on Empirical Methods in Natural Language Processing and Computational Natural Language Learning; 2007: CiteSeerx.
45. Schwartz L, Callison-Burch C, Schuler W, Wu S, editors. Incremental Syntactic Language Models for Phrase-based Translation. Proceedings of the 49th Annual Meeting of the Association for Computational Linguistics: Human Language Technologies; 2011: Association for Computational Linguistics.
46. Ding Y, Palmer M, editors. Machine translation using probabilistic synchronous dependency insertion grammars proceedings of the 43rd Annual Meeting of the ACL; 2005: Association for Computational Linguistics.
47. Weaver W. Translation. Machine translation of languages. 1949;14:15-23.
48. Nandini D. Semantic Web and Ontology-eBooks and textbooks from bookboon.com. Booknoon.com; 2014.
49. Chua CC, Lim TY, Soon L-K, Tang EK, Ranaivo-Malançon B. Meaning preservation in example-based machine translation with structural semantics. Journal of Expert Systems with Applications. 2017.
50. Shi C, Wang H, editors. Research on ontology-driven Chinese-English machine translation. Natural Language Processing and Knowledge Engineering, 2005 IEEE NLP-KE '05 Proceedings of 2005 IEEE International Conference on; 2005, 30 Oct.-1 Nov. 2005; Wuhan, China: IEEE.
51. AlAgha I, Abu-Taha AW. AR2SPARQL: An Arabic Natural Language Interface for the Semantic Web. International Journal of Computer Applications. 2015;125(6).
52. AlAgha I. Using linguistic analysis to translate Arabic natural language queries to SPARQL. arXiv preprint arXiv; 2015.
53. Abu-Taha AW, Alagha IM. An ontology-based Arabic question answering system. Central library of Islamic University of Ghaza; 2015.
54. Cho K, Merriënboer BV, Bahdanau D, Bengio Y. On the properties of neural machine translation: Encoder-decoder approaches. arXiv preprint arXiv. 2014;1409(1259).
55. Bahdanau D, Cho K, Bengio Y. Neural machine translation by jointly learning to align and translate. arXiv preprint arXiv ISO 690. 2014;1409(0473).
56. Han AL-F, Wong DF. Machine Translation Evaluation: A Survey. arXiv preprint arXiv. 2016;1605(04515).
57. Wołk K, Marasek K, editors. Neural-based machine translation for medical text domain. Based on European Medicines Agency leaflet texts. Procedia Computer Science; 2015.
58. Junczys-Dowmunt M, Dwojak T, Hoang H. Is neural machine translation ready for deployment? a case study on 30 translation directions. arXiv preprint arXiv. 2016;1610(01108).
59. Zhang J, Zong C, editors. Exploiting source-side monolingual data in neural machine translation. In Proceedings of EMNLP ISO 690; 2016.
60. Kalchbrenner N, Phil B, editors. Recurrent continuous translation models. Proceedings of the 2013 Conference on

Empirical Methods in Natural Language Processing; 2013.
61. Long Z, Mitsuhashi T, Mitsuhashi T, Yamamoto M. Translation of patent sentences with a large vocabulary of technical terms using neural machine translation. arXiv preprint arXiv. 2017;1704.04521v1 [cs.CL].
62. Sennrichy R, Firat O, Cho K, Birch A, Haddow B, Hitschler J, et al. Nematus: A toolkit for neural machine translation. arXiv preprint arXiv. 2017;1703.04357v1 [cs.CL].
63. Kalchbrenner N, Espeholt L, Simonyan K, Oord AVD, Graves A, Kavukcuoglu K. Neural Machine Translation in Linear Time. arXiv preprint arXiv. 2016;1610.10099:11.
64. Costa-Jussà MR, Fonollosa JAR. Character-based neural machine translation. arXiv preprint arXiv. 2016;1603.00810:5.
65. Kleiny G, Kim Y, Deng Y, Senellart J, Rush AM. OpenNMT: Open-source Toolkit for neural machine translation. arXiv preprint arXiv. 2017;1701.02810v2 [cs.CL].
66. Wu Y, Schuster M, Chen Z, Le QV, Norouzi M, Macherey W, et al. Google's neural machine translation system: Bridging the gap between human and machine translation. arXiv preprint arXiv. 2016;1609.08144v2:23.
67. Schmidhuber J. Deep learning in neural networks: An overview. arXiv preprint arXiv. 2014;1404.7828:88.
68. Zhou J, Cao Y, Wang X, Li P, Xu W. Deep recurrent models with fast-forward connections for neural machine translation. arXiv preprint arXiv. 2016;1606.04199v3:13.
69. Ditters E, editor. A formal grammar for the description of sentence structure in modern standard Arabic. EACL 2001 Workshop Proceedings on Arabic Language Processing: Status and Prospects; 2001.
70. Ramsay A, Mansour H. Towards including prosody in a text-to-speech system for modern standard Arabic. Computer Speech and Language. 2007;22:84-103.
71. Zabokrtsky Z, Smrž O, editors. Arabic syntactic trees: From constituency to dependency. Proceedings of the tenth conference on European chapter of the Association for Computational Linguistics, European Chapter of the Association for Computational Linguistics EACL 2003; 2003.
72. Al-Raheb Y, Akrout A, Genabith JV, Dichy J. DCU 250 Arabic dependency bank: An LFG gold standard resource for the Arabic Penn treebank. The Challenge of Arabic for NLP/MT Conference; 23 October 2006; London, UK: DCU Library. 2006;12.
73. Green S, Manning CD, editors. Better Arabic parsing: Baselines, evaluations, and analysis. Proceedings of the 23rd International Conference on Computational Linguistics, Association for Computational Linguistics; 2010 August 23 - 27, 2010 Beijing, China: Association for Computational Linguistics, ACM.
74. Tounsi L, Attia M, Genabith JV. Parsing Arabic using treebank-based LFG resources; 2009.
75. Bikel DM. Intricacies of Collins' parsing model. Computational Linguistics. 2004;30(4):479-511.
76. Zaghouani W, Habash N, Obeid O, Mohit B, Bouamor H, Oflazer K. Building an Arabic machine translation post-edited corpus: Guidelines and annotation. International Conference on Language Resources and Evaluation (LREC 2016); 2016.
77. Quirk C, Choudhury P, Gao J, Suzuki H, Toutanova K, Gamon M, et al. MSR SPLAT, a language analysis toolkit. Proceedings of the 2012 Conference of the North American Chapter of the Association for Computational Linguistics: Human Language Technologies: Demonstration Session: Association for Computational Linguistics; 2012.
78. Abdolahi M, Zahedi M. An overview on text coherence methods. Information and Knowledge Technology (IKT), 2016 Eighth International Conference on IEEE. 2016;5.
79. Soricut R, Marcu D, editors. Discourse generation using utility-trained coherence models. Proceedings of the COLING/ACL on Main conference poster sessions; 2006: Association for Computational Linguistics.
80. Alrakaf AA, Rahman SMM. A supervised approach for word sense disambiguation based on Arabic diacritics. Informatics, Electronics and Vision (ICIEV), 2016 5th International Conference on: IEEE; 2016.
81. Ye Z, Jia Z, Huang J, Yin H. Part-of-speech tagging based on dictionary and statistical machine learning. Control Conference (CCC), 2016 35th Chinese; China: IEEE; 2016.
82. Hussain S, Kuli JJ, Hazarika GC. The first step towards named entity recognition in

missing language. Electrical, Electronics, and Optimization Techniques (ICEEOT), International Conference on IEEE. 2016;4.

83. Antoniou G, Harmelen FV. A semantic web primer. MIT press; 2004.
84. Alromima W, Moawad IF, Elgohary R, Aref. M. Ontology-Based Model for Arabic Lexicons: An Application of the Place Nouns in the Holy Quran. Computer Engineering Conference (ICENCO), IEEE. 2015;11th International:7.
85. Al-Muhtaseb H, Mellish C. Some differences between Arabic and English: A step towards an Arabic upper model. 6th International Conference on Multilingual Computing1998.
86. Bella G, Giunchiglia F, McNeill F. Language and domain aware lightweight ontology matching. Web Semantics: Science, Services and Agents on the World Wide Web. 2017;43:17.
87. Zaghouani W, Awad D, editors. Building an Arabic Punctuated Corpus. Qatar Foundation Annual Research Conference Proceedings; 2016; Qatar: Qatar: HBKU Press.
88. Rafalovitch A, Dale R, editors. United Nations general assembly resolutions: A six-language parallel corpus. Proceedings of the MT Summit; 2009.
89. Gurk SM, Abela C, Debattista J. Towards ontology quality assessment. 4th Workshop on Linked Data Quality (LDQ2017); 2017.
90. Kazemi A, Toral A, Way A, Monadjemi A, Nematbakhsh M. Syntax and semantic-based reordering in hierarchical phrase-based statistical machine translation. Journal of Expert Systems With Applications. 2017;84:14.
91. Simha NP, Manohar V, Suresh MS, Bhat DD, Chakrasali S. Rule based simple english sentence correction by rearrangement of words. International Journal of Scientific & Engineering Research. 2016;7(6):3.

*Peer-review history:*
*The peer review history for this paper can be accessed here:*
*http://sciencedomain.org/review-history/20921*